\def\year{2018}\relax
\documentclass[letterpaper]{article} 
\usepackage{aaai18}  
\usepackage{times}  
\usepackage{helvet}  
\usepackage{courier}  
\usepackage{url}  
\usepackage{graphicx}  
\usepackage{subfig}
\usepackage{amsmath}
\usepackage{amsfonts}
\usepackage{amssymb}
\usepackage{amsthm}
\usepackage{bbm}
\usepackage{color}
\usepackage{subfiles}
\newtheorem{theorem}{Theorem}
\newtheorem{lemma}{Lemma}
\newtheorem*{corollary*}{Proof Sketch}

\newtheorem{proposition}{Proposition}
\usepackage[linesnumbered, algoruled]{algorithm2e}

\begin{document}
%
\title{Stochastic Non-convex Ordinal Embedding with \\Stabilized Barzilai-Borwein Step Size}
\author
{
		Ke Ma\textsuperscript{1,2}, Jinshan Zeng\textsuperscript{3,4}, Jiechao Xiong\textsuperscript{5}, Qianqian Xu\textsuperscript{1}, 
		\bf\Large{Xiaochun Cao\textsuperscript{1*}, Wei Liu\textsuperscript{5}, Yuan Yao\textsuperscript{4}\thanks{The corresponding authors.}}\\
		\textsuperscript{1} State Key Laboratory of Information Security, Institute of Information Engineering, Chinese Academy of Sciences\\
		\textsuperscript{2} School of Cyber Security, University of Chinese Academy of Sciences\\
		\textsuperscript{3} School of Computer Information Engineering, Jiangxi Normal University\\
		\textsuperscript{4} Department of Mathematics, Hong Kong University of Science and Technology \textsuperscript{5} Tencent AI Lab\\
		\{make, xuqianqian, caoxiaochun\}@iie.ac.cn, jsh.zeng@gmail.com\\
		jcxiong@tencent.com, wliu@ee.columbia.edu, yuany@ust.hk
}
\maketitle
\begin{abstract}
	Learning representation from relative similarity comparisons, often called ordinal embedding, gains rising attention in recent years. Most of the existing methods are batch methods designed mainly based on the convex optimization, say, the projected gradient descent method. However, they are generally time-consuming due to that the singular value decomposition (SVD) is commonly adopted during the update, especially when the data size is very large. To overcome this challenge, we propose a stochastic algorithm called SVRG-SBB, which has the following features: (a) SVD-free via dropping convexity, with good scalability by the use of stochastic algorithm, i.e., stochastic variance reduced gradient (SVRG), and (b) adaptive step size choice via introducing a new stabilized Barzilai-Borwein (SBB) method as the original version for convex problems might fail for the considered stochastic \textit{non-convex} optimization problem. Moreover, we show that the proposed algorithm converges to a stationary point at a rate $\mathcal{O}(\frac{1}{T})$ in our setting, where $T$ is the number of total iterations. Numerous simulations and real-world data experiments are conducted to show the effectiveness of the proposed algorithm via comparing with the state-of-the-art methods, particularly, much lower computational cost with good prediction performance.
\end{abstract}

\section{Introduction}

 Ordinal embedding aims to learn representation of data objects as points in a low-dimensional space. The distances among these points agree with a set of relative similarity comparisons as well as possible. Relative comparisons are often collected by workers who are asked to answer the following question:
	
\emph{``Is the similarity between object $i$ and $j$ larger than the similarity between $l$ and $k$?"}
	
The feedback of individuals provide us a set of quadruplets, \textit{i.e.}, $\{(i,j,l,k)\}$, which can be treated as the supervision for ordinal embedding. Without prior knowledge, the relative similarity comparisons always involve all objects and the number of potential quadruplet could be $\mathcal{O}(n^4)$.
	
The ordinal embedding problem was firstly studied by \cite{Shepard1962a,Shepard1962b,Kruskal1964a,Kruskal1964b} in the psychometric society. In recent years, it has drawn a lot of attention in machine learning \cite{jamieson2011low,53e99af7b7602d97023851bf,2015arXiv150102861A,NIPS2016_6554}, statistic ranking \cite{McFee:2011:LMS:1953048.1953063,kevin2011active}, artificial intelligence \cite{Heikinheimo2013TheCA,503}, information retrieval \cite{7410580}, and computer vision \cite{wah2014similarity,wilberKKB2015concept}, etc.

One of the typical methods for ordinal embedding problem is the well-known
Generalized Non-Metric Multidimensional Scaling (GNMDS) \cite{agarwal2007generalized}, which aims at finding a low-rank Gram (or kernel) matrix $\mathbf{G}$ in Euclidean space such that the pairwise distances between the embedding of objects in Reproducing Kernel Hilbert Space (RKHS) satisfy the relative similarity comparisons. As GNMDS uses hinge loss to model the relative similarity between the objects, it neglects the information provided by satisfied constraints in finding the underlying structure in the low-dimensional space. To alleviate this issue, the Crowd Kernel Learning (CKL) was proposed by \cite{tamuz2011adaptiive} via employing a scale-invariant loss function. However,
the objective function used in CKL
only considers the constraints which are strongly violated.
Latter, \cite{vandermaaten2012stochastic} proposed the Stochastic Triplet Embedding (STE) that jointly penalizes the violated constraints and rewards the satisfied constraints, via using the logistic loss function. Note that the aforementioned three typical methods are based on the convex formulations, and also employ the projected gradient descent method and singular value decomposition (SVD) to obtain the embedding.
However, a huge amount of comparisons and the computational complexity of SVD significantly inhibit their usage to large scale and on-line applications. Structure Preserving Embedding (SPE) \cite{Shaw:2009:SPE:1553374.1553494} and Local Ordinal Embedding (LOE) \cite{Terada2014LocalOE} embed unweighted nearest neighbor graphs to Euclidean spaces with convex and non-convex objective functions. The nearest neighbor adjacency matrix can be transformed into ordinal constraints, but it is not a standard equipment in those scenarios which involve relative comparisons. With this limitation, SPE and LOE are not suitable for ordinal embedding via quadruplets or triple comparisons.
	
In contrast to the kernel-learning or convex formulation of ordinal embedding, the aforementioned methods have the analogous non-convex counterparts. The non-convex formulations directly obtain the embedding instead of the Gram matrix. Batch gradient descent is not suitable for solving these large scale ordinal embedding problems because of the expense of full gradients in each iteration. Stochastic gradient descent (SGD) is a common technology in this situation as it takes advantage of the stochastic gradient to devise fast computation per iteration.
In \cite{ghadimi2013stochastic}, the $\mathcal{O}(\frac{1}{\sqrt{T}})$ convergence rate of SGD for the stochastic non-convex optimization problem was established, in the sense of convergence to a stationary point, where $T$ is the total number of iterations.
As SGD has slow convergence due to the inherent variance, stochastic variance reduced gradient (SVRG) method was proposed in \cite{rie2013accelerating} to accelerate SGD.
For the strongly convex function, the linear convergence of SVRG with Option-II was established in \cite{rie2013accelerating}, and latter the linear convergence rates of SVRG with Option-I and SVRG incorporating with the Barzilai-Borwein (BB) step size were shown in \cite{NIPS2016_6286}.
In the non-convex case, the $\mathcal{O}(\frac{1}{T})$ convergence rates of SVRG in the sense of convergence to a stationary point were shown in \cite{pmlr-v48-reddi16,pmlr-v48-allen-zhua16} under certain conditions.

Although the BB step size has been incorporated into SVRG and its effectiveness has been shown in
 \cite{NIPS2016_6286} for the strongly convex case, it might not work when applied to some stochastic non-convex optimization problems.
Actually, in our latter simulations, we found that the absolute value of the original BB step size is unstable when applied to the stochastic non-convex ordinal embedding problem studied in this paper (see, Figure \ref{fig:step}(a)). The absolute value of the original BB step size varies dramatically with respect to the epoch number.
Such phenomenon is mainly due to without the strong convexity, the denominator of BB step size might be very close 0, and thus the BB step size broken up. This motivates us to investigate some new stable and adaptive strategies of step size for SVRG when applied to the stochastic non-convex ordinal embedding problem.

In this paper, we introduce a new adaptive step size strategy called stabilized BB (SBB) step size via adding another positive term to the absolute of the denominator of the original BB step size to overcome the instability of the BB step size, and then propose a new stochastic algorithm called SVRG-SBB via incorporating the SBB step size for fast solving the considered non-convex ordinal embedding model. In a summary, our main contribution can be shown as follows:	
\begin{itemize}
\item
We propose a non-convex framework for the ordinal embedding problem via considering the optimization problem with respect to the original embedding variable but not its Gram matrix. By exploiting this idea, we get rid of the positive semi-definite (PSD) constraint on the Gram matrix, and thus, our proposed algorithm is SVD-free and has good scalability. 

\item
The introduced SBB step size can overcome the instability of the original BB step size when the original BB step size does not work. More importantly, the proposed SVRG-SBB algorithm outperforms most of the the state-of-the-art methods as shown by numerous simulations and real-world data experiments, in the sense that SVRG-SBB often has better generalization performance and significantly reduces the computational cost.

\item We establish $O(\frac{1}{T})$ convergence rate of SVRG-SBB in the sense of convergence to a stationary point, where $T$ is the total number of iterations. Such convergence result is comparable with the existing best convergence results in literature.
\end{itemize}

\section{Stochastic Ordinal Embedding}

\subsection{A. Problem Description}

 There is a set of $n$ objects $\{o_1,\dots,o_n\}$ in abstract space $\mathbf{O}$. We assume that a certain but unknown dissimilarity function $\xi:\mathbf{O}\times\mathbf{O}\rightarrow\mathbb{R}^{+}$ assigns the dissimilarity value $\xi_{ij}$ for a pair of objects $(o_i,o_j)$. With the dissimilarity function $\xi$, we can define the ordinal constraint $(i,j,l,k)$ from a set $\mathcal{P}\subset[n]^4$, where
$$
	\mathcal{P}=\{(i,j,l,k)\ |\ \text{if exist }o_i,o_j,o_k,o_l\text{ satisfy }\xi_{ij}<\xi_{lk}\}
$$
and $[n]$ is the set of $\{1,\dots,n\}$. Our goal is to obtain the representations of $\{o_1,\dots,o_n\}$ in Euclidean space $\mathbb{R}^{p}$ where $p$ is the desired embedding dimension. The embedding $\mathbf{X}\in\mathbb{R}^{n\times d}$ should preserve the ordinal constraints in $\mathcal{P}$ as much as possible, which means
$$0
	(i,j,l,k)\in\mathcal{P} \Leftrightarrow \xi_{ij} < \xi_{lk} \Leftrightarrow d^2_{ij}(\mathbf{X}) < d^2_{lk}(\mathbf{X})
$$
where $d^2_{ij}(\mathbf{X})=\|\mathbf{x}_i-\mathbf{x}_j\|^2$ is the squared Euclidean distance between $\mathbf{x}_i$ and $\mathbf{x}_j$, and $\mathbf{x}_i$ is the $i^{th}$ row of $\mathbf{X}$.
	
Let $\mathbf{D}=\{d^2_{ij}(\mathbf{X})\}$ be the distance matrix of $\mathbf{X}$. There are some existing methods for recovering $\mathbf{X}$ given ordinal constraints on distance matrix $\mathbf{D}$. It is known that $\mathbf{D}$ can be determined by the Gram matrix $\mathbf{G} = \mathbf{X}\mathbf{X}^T = \{g_{ij}\}^{n}_{i,j=1}$ as
$
	d^2_{ij}(\mathbf{X}) = g_{ii}-2g_{ij}+g_{jj},
$
and
$$
	\mathbf{D} = \textit{diag}(\mathbf{G})\cdot\mathbf{1}^T-2\mathbf{G}+\mathbf{1}\cdot\textit{diag}(\mathbf{G})^T
$$
where $\textit{diag}(\mathbf{G})$ is the column vector composed of the diagonal of $\mathbf{G}$ and $\mathbf{1}^T=[1,\dots,1]$. As $\text{rank}(\mathbf{G})\leq\min(n, d)$ and it always holds $d\ll n$, these methods \cite{agarwal2007generalized,tamuz2011adaptiive,vandermaaten2012stochastic} can be generalized by a semidefinite programming (SDP) with low rank constraint,
\begin{equation}
	\label{eq:1}
		\underset{\mathbf{G}\in\mathbb{R}^{n\times n}}{\min} \ \ L(\mathbf{G})+\lambda\cdot\text{tr}(\mathbf{G}) \quad
		\text{s.t.} \quad \mathbf{G}\succeq 0
\end{equation}
where
$
	L(\mathbf{G})=\frac{1}{|\mathcal{P}|}\sum_{p\in\mathcal{P}}l_p(\mathbf{G})
$
is a convex function of $\mathbf{G}$ which satisfies
\begin{equation*}
	l_p(\mathbf{G}):
	\left\{
	\begin{matrix}
	> 0,\ & d^2_{ij}(\mathbf{X}) > d^2_{lk}(\mathbf{X})\\
	\leq 0,\ & \text{otherwise.}
	\end{matrix}
	\right.
\end{equation*}
$\text{tr}(\mathbf{G})$ is the trace of matrix $\mathbf{G}$. To obtain the embedding $\mathbf{X}\in\mathbb{R}^{n\times d}$, the projected gradient descent is performed. The basic idea of the projected gradient descent method is: the batch gradient descent step with all $p\in\mathcal{P}$ is firstly used to learn the Gram matrix $\mathbf{G}$,
$$
	{\mathbf{G}}'_{t} = \mathbf{G}_{t-1}-\eta_{t}(\nabla L(\mathbf{G}_{t-1})+\lambda \mathbf{I})
$$
where $t$ denotes the current iteration, $\eta_t$ is the step size; then ${\mathbf{G}}'_{t}$ is projected onto a positive semi-definite (PSD) cone $\mathbb{S}_+$,
$
	\mathbf{G}_{t} = \Pi_{\mathbb{S}_+}({\mathbf{G}}'_{t});
$
and latter, once the iterates converge, the embedding $X$ is obtained by projecting $\mathbf{G}$ onto the subspace spanned by the largest $p$ eigenvectors of $\mathbf{G}$ via SVD.

\subsection{B. Stochastic Non-convex Ordinal Embedding}

Although the SDP (\ref{eq:1}) is a convex optimization problem, there are some disadvantages of this approach: (i) the projection onto PSD cone $\mathbb{S}_+$, which is performed by an expensive SVD due to the absence of any prior knowledge on the structure of $\mathbf{G}$, is a computational bottleneck of optimization; and (ii) the desired dimension of the embedding $\mathbf{X}$ is $d$ and we hope the Gram matrix $\mathbf{G}$ satisfies $\text{rank}(\mathbf{G})\leq d$. If $\text{rank}(\mathbf{G})\gg d$, the freedom degree of $\mathbf{G}$ is much larger than $\mathbf{X}$ with over-fitting. Although $\mathbf{G}$ is a global optimal solution of (\ref{eq:1}), the subspace spanned by the largest $d$ eigenvectors of $\mathbf{G}$ will produce less accurate embedding. We can tune the regularization parameter $\lambda$ to force $\{\mathbf{G}_t\}$ to be low-rank and cross-validation is the most utilized technology. This needs extra computational cost. In summary, projection and parameter tuning render gradient descent methods computationally prohibitive for learning the embedding $\mathbf{X}$ with ordinal information $\mathcal{P}$. To overcome these challenges, we will exploit the non-convex and stochastic optimization techniques for the ordinal embedding problem.
	
	
To avoid projecting the Gram matrix $\mathbf{G}$ onto the PSD cone $\mathbb{S}_{+}$ and tuning the parameter $\lambda$, we directly optimize $\mathbf{X}$ and propose the unconstrained optimization problem of learning embedding $\mathbf{X}$,
\begin{equation}
	\label{eq:2}
	\underset{\mathbf{X}\in\mathbb{R}^{n\times d}}{\min}\ F(\mathbf{X}):=\frac{1}{|\mathcal{P}|}\ \underset{p\in\mathcal{P}}{\sum}\ f_p(\mathbf{X})
\end{equation}
where $f_p(\mathbf{X})$ evaluates
$$
	\triangle_p = d^2_{ij}(\mathbf{X})-d^2_{lk}(\mathbf{X}),\ p=(i,j,l,k).
$$
and
\begin{equation*}
	f_p(\mathbf{X}):
	\left\{
	\begin{matrix}
	\leq 0,\ & \triangle_p\leq 0\\
	> 0,\ & \text{otherwise.}
	\end{matrix}
	\right.
\end{equation*}
The loss function $f_p(\mathbf{X})$ can be chosen as the hinge loss \cite{agarwal2007generalized}
\begin{equation}
	\label{eq:hinge}
	f_p(\mathbf{X}) = \max\{0, 1+\triangle_p\},
\end{equation}
the scale-invariant loss \cite{tamuz2011adaptiive}
\begin{equation}
	\label{eq:scale-invariant}
	f_p(\mathbf{X}) = \log\frac{d^2_{lk}(\mathbf{X})+\delta}{d^2_{ij}(\mathbf{X})+d^2_{lk}(\mathbf{X})+2\delta},
\end{equation}
where $\delta\neq 0$ is a scalar which overcomes the problem of degeneracy and preserve numerical stable, the logistic loss \cite{vandermaaten2012stochastic}
\begin{equation}
	\label{eq:logistic}
	f_p(\mathbf{X}) = -\log(1+\exp(\triangle_p)),
\end{equation}
and replacing the Gaussian kernel in (\ref{eq:logistic}) by the Student-$t$ kernel with degree $\alpha$ \cite{vandermaaten2012stochastic}
\begin{equation}
	\label{eq:student}
	f_p(\mathbf{X}) = -\log\frac{\left(1+\frac{d^2_{ij}(\mathbf{X})}{\alpha}\right)^{-\frac{\alpha+1}{2}}}{\left(1+\frac{d^2_{ij}(\mathbf{X})}{\alpha}\right)^{-\frac{\alpha+1}{2}}+\left(1+\frac{d^2_{lk}(\mathbf{X})}{\alpha}\right)^{-\frac{\alpha+1}{2}}}.
\end{equation}
Since \eqref{eq:2} is an unconstrained optimization problem, it is obvious that SVD and parameter $\lambda$	are avoided. Moreover, instead of the batch methods like the gradient descent method, we use a fast stochastic gradient descent algorithm like SVRG to solve the non-convex problem \eqref{eq:2}.

\section{SVRG with Stabilized BB Step Size}


\subsection{A. Motivation}

\begin{figure}[!t]
	\centering
	\subfloat[{SBB$_0$ step size}]
	{
		\includegraphics[width = 0.4\columnwidth]{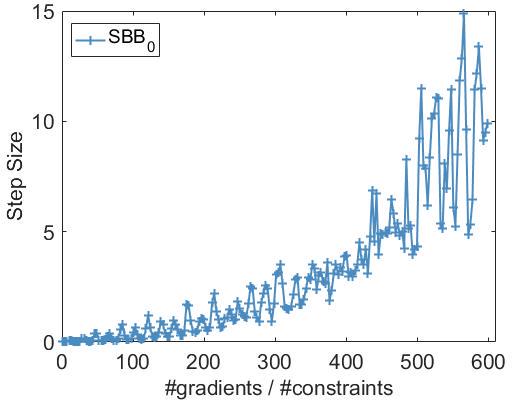}
		\label{fig:syth:tste:bb:step}
	}
	\subfloat[{SBB$_{0.005}$ step size}]
	{
		\includegraphics[width = 0.4\columnwidth]{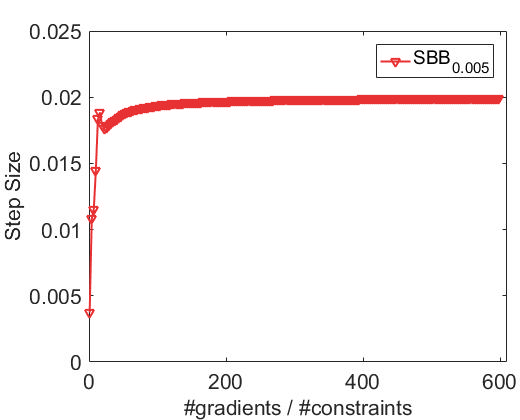}
		\label{fig:syth:tste:rbb:step}
	}
	\caption{Step sizes along iterations of SVRG-SBB$_\epsilon$ on the synthetic data. (a) SBB step size with $\epsilon=0$, (b) SBB step size with $\epsilon=0.005$.}
	\label{fig:step}
\end{figure}

One open issue in stochastic optimization is how to choose an appropriate step size for SVRG in practice. The common method is either to use a constant step size to track, a diminishing step size to enforce convergence, or to tune a step size empirically, which can be time consuming. Recently, \cite{NIPS2016_6286} proposed to use the Barzilai-Borwein (BB) method to automatically compute step sizes in SVRG for strongly convex objective function shown as follows
\begin{equation}
	\label{eq:bb_step}
	\eta_{s} = \frac{1}{m}\frac{\|\tilde{\mathbf{X}}^{s}-\tilde{\mathbf{X}}^{s-1}\|^2_F}{\text{vec}(\tilde{\mathbf{X}}^{s}-\tilde{\mathbf{X}}^{s-1})^T\text{vec}(\mathbf{g}^s-\mathbf{g}^{s-1})},
\end{equation}
where $\tilde{\mathbf{X}}^{s}$ is the $s$-th iterate of the outer loop of SVRG and $\mathbf{g}^{s} = \nabla F(\tilde{\mathbf{X}}^s)$.
However, if the objective function $F$ is non-convex, the denominator of \eqref{eq:bb_step} might be close to 0 and even negative that fail the BB method. For example, Figure \ref{fig:step}(a) shows that in simulations one can observe the instability of the absolute value of the original BB step size (called SBB$_0$ henceforth) in non-convex problems. Due to this issue, the original BB step size might not be suitable for the non-convex ordinal embedding problem.

%
%

\subsection{B. Stabilized BB step size}

An intuitive way to overcome the flaw of BB step size is to add another positive term in the absolute value of the denominator of the original BB step size, which leads to our introduced stabilized Barzilai-Borwein (SBB) step size shown as follows,
\begin{equation}
	\label{eq:rbb_step}
	\begin{aligned}
		& \eta_{s} &=&\ \ \frac{1}{m}\cdot\left\|\tilde{\mathbf{X}}^{s}-\tilde{\mathbf{X}}^{s-1}\right\|^2_F\\
		& &\times& \ \ \left(\left|\text{vec}(\tilde{\mathbf{X}}^{s}-\tilde{\mathbf{X}}^{s-1})^T\text{vec}(\mathbf{g}^s-\mathbf{g}^{s-1})\right|\right.\\
		& &+&\ \ \left.\epsilon\left\|\tilde{\mathbf{X}}^{s}-\tilde{\mathbf{X}}^{s-1}\right\|^2_F\right)^{-1}, \quad \text{for some} \  \epsilon>0.
	\end{aligned}
\end{equation}
By the use of SBB step size, the SVRG-SBB algorithm is presented in Algorithm \ref{alg:svrg-bb}.

Actually, as shown by our latter theorem (i.e., Theorem \ref{svrg_bb_nonconvex}), if the Hessian of the objective function $\nabla^2 F(X)$ is nonsingular and the magnitudes of its eigenvalues are lower bounded by some positive constant $\mu$, then we can take $\epsilon=0$. In this case, we call the referred step size SBB$_0$ henceforth. Even if we have no information of the Hessian of the objective function in practice, the SBB$_\epsilon$ step size with an $\epsilon>0$ is just a more consecutive step size of SBB$_0$ step size.

From \eqref{eq:rbb_step}, if the gradient $\nabla F$ is Lipschitz continuous with constant $L>0$, then the SBB$_\epsilon$ step size can be bounded as follows
\begin{align}
\label{eq:bound-rbb}
\frac{1}{m(L+\epsilon)} \leq \eta_k \leq \frac{1}{m\epsilon},
\end{align}
where the lower bound is obtained by the $L$-Lipschitz continuity of $\nabla F$, and the upper bound is directly derived by its specific form.
If further $\nabla^2 F(X)$ is nonsingular and the magnitudes of its eigenvalues has a lower bound $\mu>0$,
then the bound of SBB$_0$ becomes
\begin{align}
\label{eq:bound-rbb0}
\frac{1}{m L} \leq \eta_k \leq \frac{1}{m \mu}.
\end{align}

As shown in Figure \ref{fig:step} (b), SBB$_\epsilon$ step size with a positive $\epsilon$ can make SBB$_0$ step size more stable when SBB$_0$ step size is unstable and varies dramatically. Moreover, SBB$_\epsilon$ step size usually changes significantly only at the initial several epoches, and then quickly gets very stable. This is mainly because there are many iterations in an epoch of SVRG-SBB, and thus, the algorithm might close to a stationary point after only one epoch, and starting from the second epoch, the SBB$_\epsilon$ step sizes might be very close to the inverse of the sum of the curvature of objective function and the parameter $\epsilon$ used.

\begin{algorithm}[ht]
	\caption{SVRG-SBB for \eqref{eq:2}}
	\label{alg:svrg-bb}
	\KwIn{$\scriptstyle\epsilon\geq0$, update frequency $m$, maximal number of iterations $S$, initial step size $\scriptstyle\eta_0$ (only used in the first epoch), initial point $\scriptstyle\tilde{\bf X}^0 \in \mathbb{R}^{n\times d}$, and $\scriptstyle N: = |{\cal P}|$}
	\KwOut{${\scriptstyle \mathbf{X}_{\mathrm{out}}}$ is chosen uniformly from $\scriptstyle\{\{\mathbf{X}^{s}_{t}\}_{t=1}^{m}\}_{s=1}^{S}$}
	\For{$\scriptstyle s=0$ to $\scriptstyle S-1$}
	{
		$\scriptstyle \mathbf{g}^{s} = \nabla F(\tilde{\mathbf{X}}^s) =\frac{1}{N}\sum^N_{i=1}\nabla f_i(\tilde{\mathbf{X}}^s)$\;
		\If{$\scriptstyle s>0$}
		{
			\begin{equation}
				\scriptstyle
				\tag{\ref{eq:rbb_step}}
				\eta_{s} = \frac{1}{m}\cdot\frac{\|\tilde{\mathbf{X}}^{s}-\tilde{\mathbf{X}}^{s-1}\|^2_F}{|\text{vec}(\tilde{\mathbf{X}}^{s}-\tilde{\mathbf{X}}^{s-1})^T\text{vec}(\mathbf{g}_s-\mathbf{g}_{s-1})|+\epsilon\|\tilde{\mathbf{X}}^{s}-\tilde{\mathbf{X}}^{s-1}\|^2_F}\;
			\end{equation}
		}
		$\scriptstyle\mathbf{X}^{s}_0=\tilde{\mathbf{X}}^s$\;
		\For{$ \scriptstyle t=0$ to $\scriptstyle m-1$}
		{
			uniformly randomly pick $\scriptstyle i_t\in \{1,\ldots,N\}$\;
			$\scriptstyle\mathbf{X}^{s}_{t+1}= \mathbf{X}^{s}_{t}-\eta_{s}(\nabla f_{i_t}(\mathbf{X}^{s}_{t})-\nabla f_{i_t}(\tilde{\mathbf{X}}^s)+\mathbf{g}^{s})$\;
		}
		$\scriptstyle \tilde{\mathbf{X}}^{s+1}=\mathbf{X}^{s}_{m}$\;
	}
\end{algorithm}

\subsection{C. Convergence Results}

In this subsection, we establish the convergence rate of SVRG-SBB as shown in the following theorem.

\begin{theorem}
\label{svrg_bb_nonconvex}
Let $\{\{{\bf X}_t^s\}_{t=1}^m\}_{s=1}^S$ be a sequence generated by Algorithm \ref{alg:svrg-bb}.
Suppose that $F$ is smooth, and $\nabla F$ is Lipschitz continuous with Lipschitz constant $L>0$ and bounded. For any $\epsilon>0$, if
\begin{align}
\label{Eq:cond-m}
m > \max \left\{ \frac{L^2}{\epsilon}\left(1+\frac{2L}{\epsilon}\right), 1+ \sqrt{1+\frac{8L^3}{\epsilon}}\right\} \cdot \epsilon^{-1},
\end{align}
then for the output ${\bf X}_{\mathrm{out}}$ of Algorithm \ref{alg:svrg-bb}, we have
\begin{align}
\label{eq:rate}
\mathbb{E}[\|\nabla F({\bf X}_{\mathrm{out}})\|^2] \leq \frac{F(\tilde{\bf X}^0)-F({\bf X}^*)}{T \cdot \gamma_S},
\end{align}
where ${\bf X}^*$ is an optimal solution of \eqref{eq:2}, $T = m \cdot S$ is the total number of iterations, $\gamma_S$ is some positive constant satisfying
\[
\gamma_S \geq \min_{0\leq s \leq S-1} \left\{ \eta_s \left[ \frac{1}{2} - \eta_s\left(1+4(m-1)L^3\eta_s^2\right)\right]\right\},
\]
and $\{\eta_s\}_{s=0}^{S-1}$ are SBB step sizes specified in \eqref{eq:rbb_step}.

If further the Hessian $\nabla^2 F(X)$ exists and $\mu$ is the lower bound of the magnitudes of eigenvalues of $\nabla^2 F(X)$ for any bounded $X$, then the convergence rate \eqref{eq:rate} still holds for SVRR-SBB  with $\epsilon$ replaced by $\mu+\epsilon$. In addition, if $\mu>0$, then we can take $\epsilon=0$, and \eqref{eq:rate} still holds for SVRR-SBB$_0$ with $\epsilon$ replaced by $\mu$.
\end{theorem}

Theorem \ref{svrg_bb_nonconvex} is an adaptation of \cite[Theorem 2]{pmlr-v48-reddi16} via noting that the used SBB step size specified in \eqref{eq:rbb_step} satisfies \eqref{eq:bound-rbb}.
The proof of this theorem is presented in supplementary material. Theorem~\ref{svrg_bb_nonconvex} shows certain non-asymptotic rate of convergence of the Algorithm~\ref{alg:svrg-bb} in the sense of convergence to a stationary point.
Similar convergence rates of SVRG under different settings have been also shown in \cite{pmlr-v48-reddi16,pmlr-v48-allen-zhua16}.

Note that the Lipschitz differentiability of the objective function is crucial for the establishment of the convergence rate of SVRG-SBB in Theorem \ref{svrg_bb_nonconvex}. In the following, we give a lemma to show that a part of aforementioned objective functions (\ref{eq:scale-invariant}), (\ref{eq:logistic}) and (\ref{eq:student}) in the ordinal embedding problem are Lipschitz differentiable. Considering the limited space of this paper, the readers are refered to (\url{}) for detailed proofs.
	
\begin{lemma}
\label{lemma:1}
The ordinal embedding functions (\ref{eq:scale-invariant}), (\ref{eq:logistic}) and (\ref{eq:student}) are Lipschitz differentiable for any bounded variable $X$.
\end{lemma}

\section{Experiments}
\label{section:experiment}
 In this section, we conduct a series of simulations and real-world data experiments to demonstrate the effectiveness of the proposed algorithms.
 Three models including \textit{GNMDS}, \textit{STE} and \textit{TSTE} are taken into consideration. Our source code could be found on the web\footnote{\url{https://github.com/alphaprime/Stabilized_Stochastic_BB}}.
	
\subsection{A. Simulations}

{
	\begin{table}[tbh!]
		\centering
		\caption{Computational complexity (second) comparisons on the synthetic dataset.}
		\resizebox{0.9\columnwidth}{!}
		{
			\begin{tabular}{c||cccc}
				\hline
				\multicolumn{5}{c}{GNMDS} \\ \hline\hline
				                      & min     & mean   & max    & std     \\ \hline
				  cvx              & 1.6140  & 1.7881 & 1.9390 & 0.0844  \\ \hline
				  ncvx Batch    & 4.5070  & 4.9372 & 5.2910 & 0.1857  \\ \hline
				  ncvx SGD      & 5.3070  & 5.5882 & 5.8970 & 0.1216  \\ \hline
				  ncvx SVRG     & 2.3020  & 2.8919 & 3.5280 & 0.2911  \\ \hline
				  ncvx SVRG-SBB$_0$  & \textbf{0.3500}  & \textbf{0.4347} & \textbf{0.5340} & \textbf{0.0367}  \\ \hline
				  ncvx SVRG-SBB$_\epsilon$ & \textit{0.3570}  & \textit{0.4858} & \textit{0.6070} & \textit{0.0621}  \\
				\hline
				\multicolumn{5}{c}{STE} \\ \hline\hline
				                    & min     & mean   & max    & std     \\ \hline
				cvx             & 3.6740  & 3.9442 & 4.1870 & 0.1709  \\ \hline
				ncvx Batch    & 2.4610  & 2.6326 & 2.9110 & 0.1162  \\ \hline
				ncvx SGD      & 1.5000  & 2.1312 & 2.7190 & 0.2740  \\ \hline
				ncvx SVRG     & 1.9930  & 2.4068 & 2.8350 & 0.1935  \\ \hline
				ncvx SVRG-SBB$_0$  & \textit{0.5000}  & \textit{0.6052} & \textit{0.6980} & \textit{0.0660}  \\ \hline
				ncvx SVRG-SBB$_\epsilon$ & \textbf{0.4510}  & \textbf{0.5773} & \textbf{0.6780} & \textbf{0.0515} \\
				\hline
				\multicolumn{5}{c}{TSTE} \\ \hline\hline
				                    & min     & mean   & max     & std    \\ \hline
				ncvx Batch    & 3.9380  & 4.1815 & 4.4790  & 0.1146 \\ \hline
				ncvx SGD      & 6.0410  & 8.2870 & 9.4770  & 0.6863 \\ \hline
				ncvx SVRG     & 1.6090  & 1.9250 & 2.3470  & 0.1807 \\ \hline
				ncvx SVRG-SBB$_0$  & \textit{0.4580}  & \textit{0.7906} & \textit{1.2480}  & \textit{0.1969} \\ \hline
				ncvx SVRG-SBB$_\epsilon$ & \textbf{0.3800}  & \textbf{0.4726} & \textbf{0.5470}  & \textbf{0.0420} \\
				\hline
			\end{tabular}
		}
		\label{tabl:1}
	\end{table}
}

\begin{figure*}[thb!]
	\centering
	\subfloat[GNMDS]
	{
		\includegraphics[width = 0.3\textwidth]{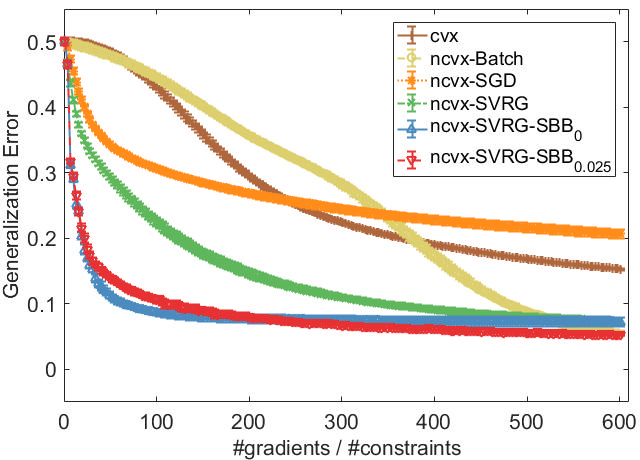}
		\label{fig:syth:gnmds:error}
	}
	\subfloat[STE]
	{
		\includegraphics[width = 0.3\textwidth]{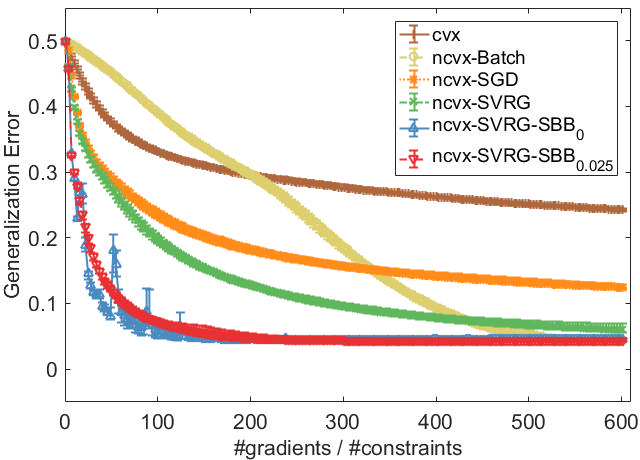}
		\label{fig:syth:ste:error}
	}
	\subfloat[TSTE]
	{
		\includegraphics[width = 0.3\textwidth]{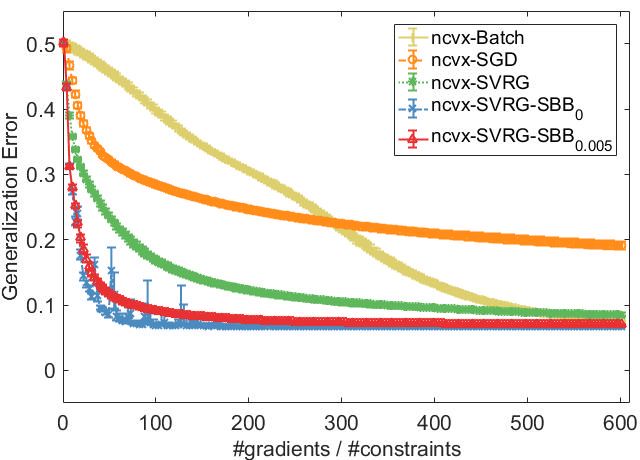}
		\label{fig:syth:tste:error}
	}
	\caption{Generalization errors of SGD, SVRG,  SVRG-SBB and batch methods on the synthetic dataset.}
	\label{fig:synthetic}
\end{figure*}

We start with a small-scale synthetic experiment to show how the methods perform in an idealized setting, which provides sufficient ordinal information in noiseless case.
\\\\
\textbf{Settings.} The synthesized dataset consists of $100$ points $\{\mathbf{x}_i\}_{i=1}^{100}\subset\mathbb{R}^{10}$, where $\mathbf{x}_i\sim\mathcal{N}(\mathbf{0}, \frac{1}{20}\mathbf{I})$, where $\mathbf{I}\in \mathbb{R}^{10\times 10}$ is the identity matrix. The possible similarity triple comparisons are generated based on the Euclidean distances between $\{\mathbf{x}_i\}$. As \cite{NIPS2016_6554} has proved that the Gram matrix $\mathbf{G}$ can be recovered from $\mathcal{O}(pn\log n)$ triplets, we randomly choose $|\mathcal{P}|=10,000$ triplets as the training set and the rest as test set. The regularization parameter and step size settings for the convex formulation follow the default setting of the STE/TSTE implementation\footnote{\url{http://homepage.tudelft.nl/19j49/ste/Stochastic_Triplet_Embedding.html}}, so we do not choose the step size by line search or the halving heuristic for convex formulation. The embedding dimension is selected just to be equal to $10$ without variations, because the results of different embedding dimensions have been discussed in the original papers of GNMDS, STE and TSTE.
\\\\
\textbf{Evaluation Metrics. }The metrics that we used in the evaluation of various algorithms include the generalization error and running time. As the learned embedding $\mathbf{X}$ from partial triple comparisons set $\mathcal{P}\subset[n]^3$ may be generalized to unknown triplets, the percentage of held-out triplets which is satisfied in the embedding $\mathbf{X}$ is used as the main metric for evaluating the quality. The running time is the duration of a algorithm when the training error is larger than $0.15$.
\\\\
\textbf{Competitors.} We evaluate both convex and non-convex formulations of three objective functions (i.e. GNMDS, STE and TSTE). We set the two baselines as : ($1$) the convex objective function whose results are denoted as ``convex'', and ($2$) these non-convex objective functions solved by batch gradient descent denoted as ``ncvx batch''. We compare the performance of SVRG-SBB$_\epsilon$ with SGD, SVRG with a fixed step size (called SVRG for short henceforth) as well as the batch gradient descent methods.
As SVRG and its variant (SVRG-SBB$_\epsilon$) runs $2m+|\mathcal{P}|$ times of (sub)gradient in each epoch, the batch and SGD solutions are evaluated with the same numbers of (sub)gradient of SVRG. In Figure~\ref{fig:synthetic}, the $x$-axis is the computational cost measured by the number of gradient evaluations divided by the total number of triple-wise constraints $|\mathcal{P}|$. The generalization error is the result of $50$ trials with different initial $\mathbf{X}_0$. For each epoch, the median number of generalization error over 50 trials with [0.25, 0.75] confidence interval are plotted. The experiment results are shown in Figure \ref{fig:synthetic} and Table \ref{tabl:1}.
\\\\
\textbf{Results.}
From to Figure \ref{fig:synthetic}, the following phenomena can be observed.
First, the algorithm SVRG-SBB$_0$ will be unstable at the initial several epoches for three models, and latter get very stable. The eventual performance of SVRG-SBB$_0$ and that of SVRG-SBB$_\epsilon$ are almost the same in three cases.
Second, compared to the batch methods, all the stochastic methods including SGD, SVRG and SVRG-SBB$_{\epsilon}$ ($\epsilon =0$ or $\epsilon>0$) converge fast at the initial several epoches and quickly get admissible results with relatively small generalization error. This is one of our main motivations to use the stochastic methods. Particularly, for all three models, SVRG-SBB$_\epsilon$ outperforms all the other methods in the sense that it not only converges fastest but also achieves almost the best generalization error. Moreover, the outperformance of SVRG-SBB$_\epsilon$ in terms of the cpu time can be also observed from Table \ref{tabl:1}. Specifically, the speedup of SVRG-SBB$_\epsilon$ over SVRG is about 4 times for all three models.


%

Table \ref{tabl:1} shows the computational complexity achieved by SGD, SVRR-SBB$_\epsilon$ and batch gradient descent for convex and non-convex objective functions. All computation is done using MATLAB$^\text{\textregistered{}}$ R2016b, on a desktop PC with Windows$^\text{\textregistered{}}$ $7$ SP$1$ $64$ bit, with $3.3$ GHz Intel$^\text{\textregistered{}}$ Xeon$^\text{\textregistered{}}$ E3-1226 v3 CPU, and $32$ GB $1600$ MHz DDR3 memory. It is easy to see that for all objective functions, SVRG-SBB$_\epsilon$ gains speed-up compared to the other methods. Besides, we notice that the convex methods could be effective when $n$ is small as the projection operator will not be the bottleneck of the convex algorithm.
{
	\begin{table*}[tbh]
		\centering
		\caption{Image retrieval performance (MAP and Precision@40) on SUN397 when $p=19$.}
		{
			\begin{tabular}{c||cc||cc||cc}
				\hline
				& \multicolumn{2}{c||}{$0\%$} & \multicolumn{2}{c||}{$5\%$} & \multicolumn{2}{c}{$10\%$}\\
				&  MAP& Precision@40 & MAP 	& Precision@40 & MAP & Precision@40 \\ \hline\hline
				\multicolumn{7}{c}{GNMDS} \\\hline\hline
				cvx              & 0.2691  & 0.3840  & 0.2512  & 0.3686   & 0.2701  & 0.3883  \\ \hline
				ncvx Batch    & 0.3357  & 0.4492  & 0.3791  & 0.4914   & 0.3835  & 0.4925  \\ \hline
				ncvx SGD      & 0.3245  & 0.4379  & 0.3635  & 0.4772   & 0.3819  & 0.4931  \\ \hline
				ncvx SVRG     & 0.3348  & 0.4490  & \textit{0.3872}  & \textit{0.4974}   & \textit{0.3870}  & \textit{0.4965}  \\ \hline
				ncvx SVRG-SBB$_0$  & \textbf{0.3941}  & \textbf{0.5040}  & 0.3700  & 0.4836   & 0.3550  & 0.4689  \\ \hline
				ncvx SVRG-SBB$_\epsilon$ & \textit{0.3363}  & \textit{0.4500}  & \textbf{0.3887}  & \textbf{0.4981}   & \textbf{0.3873}  & \textbf{0.4987} \\
				\hline\hline
				\multicolumn{7}{c}{STE} \\\hline\hline
				cvx              & 0.2114  & 0.3275  & 0.1776 & 0.2889 & 0.1989 & 0.3190  \\ \hline
				ncvx Batch    & 0.2340  & 0.3525  & 0.2252 & 0.3380 & 0.2297 & 0.3423  \\ \hline
				ncvx SGD      & 0.3369  & 0.4491  & 0.2951 & 0.4125 & 0.2390 & 0.3488  \\ \hline
				ncvx SVRG     & 0.3817  & 0.4927  & 0.3654 & 0.4804 & 0.3245 & 0.4395  \\ \hline
				ncvx SVRG-SBB$_0$  & \textbf{0.3968}   & \textbf{0.5059} & \textbf{0.3958} & \textbf{0.5054} & \textit{0.3895} & \textbf{0.5002}  \\ \hline
				ncvx SVRG-SBB$_\epsilon$ & \textit{0.3940}  & \textit{0.5036}  & \textit{0.3921} & \textit{0.5012} & \textbf{0.3896} & \textit{0.4992}  \\
				\hline\hline
				\multicolumn{7}{c}{TSTE} \\\hline\hline
				ncvx Batch    & 0.2268  & 0.3470  & 0.2069  & 0.3201 & 0.2275  & 0.3447  \\ \hline
				ncvx SGD      & 0.2602  & 0.3778  & 0.2279  & 0.3415 & 0.2402  & 0.3514  \\ \hline
				ncvx SVRG     & 0.3481  & 0.4617  & 0.3160  & 0.4332 & 0.2493  & 0.3656  \\ \hline
				SVRG-SBB$_0$             &\textbf{0.3900}    & \textbf{0.4980}  & \textbf{0.3917}  & \textbf{0.5018} & \textbf{0.3914}  & \textit{0.5007}  \\ \hline
				ncvx SVRG-SBB$_\epsilon$ & \textit{0.3625}  & \textit{0.4719}  & \textit{0.3845}  & \textit{0.4936} & \textit{0.3897}  & \textbf{0.5013} \\
				\hline\hline
			\end{tabular}
		}
		\label{tabl:3}
	\end{table*}
}

\begin{figure*}[thb!]
	\centering
	\subfloat[GNMDS]
	{
		\includegraphics[width=0.3\textwidth]{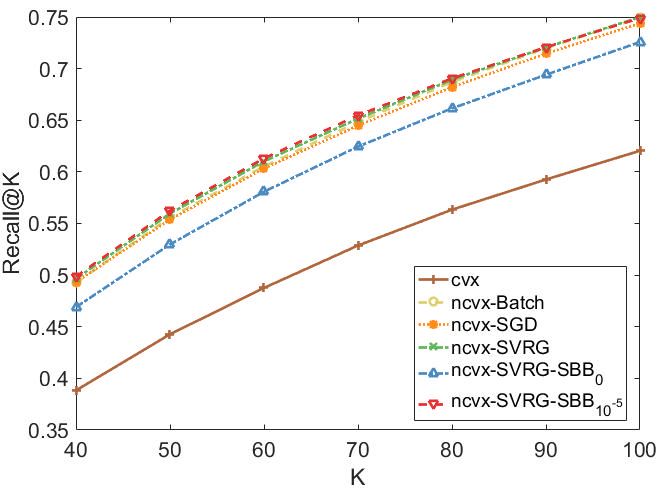}
		\label{fig:imagenet:gnmds}
	}
	\subfloat[STE]
	{
		\includegraphics[width=0.3\textwidth]{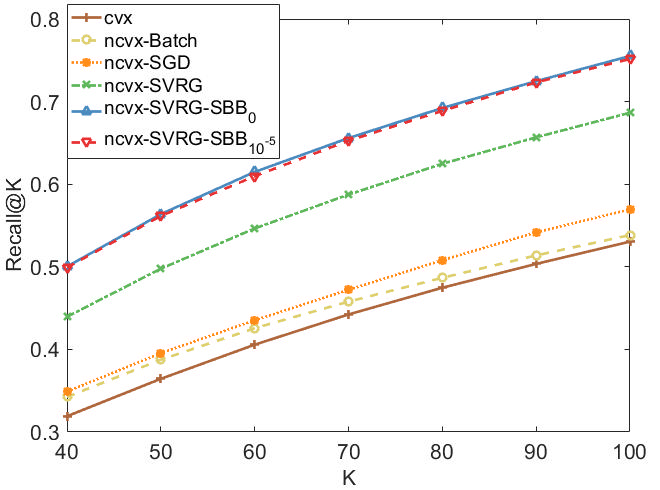}
		\label{fig:imagenet:ste}
	}
	\subfloat[TSTE]
	{
		\includegraphics[width=0.3\textwidth]{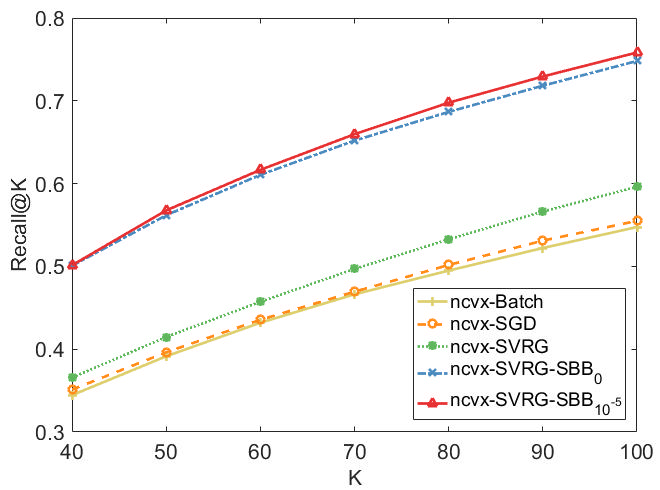}
		\label{fig:imagenet:tste}
	}
	\caption{Recall@K with $10\%$ noise on SUN397.}
	\label{fig:sun} 
\end{figure*}

\subsection{B. Image Retrieval on SUN397}

Here, we apply the proposed SVRG-SBB algorithm for a real-world dataset, i.e., SUN 397, which is generally used for the image retrieval task.
In this experiment, we wish to see how the learned representation characterizes the ``relevance'' of the same image category and the ``discrimination'' of different image categories. Hence, we use the image representation obtained by ordinal embedding for image retrieval.
\\\\
\textbf{Settings. }We evaluate the effectiveness of the proposed stochastic non-convex ordinal embedding method for visual search on the SUN397 dataset. SUN$397$ consists of around 108K images from $397$ scene categories. In SUN$397$, each image is represented by a $1,600$-dimensional feature vector extracted by principle component analysis (PCA) from $12,288$-dimensional Deep Convolution Activation Features \cite{Gong2014}. We form the training set by randomly sampling $1,080$ images from $18$ categories with $60$ images in each category. Only the training set is used for learning an ordinal embedding and a nonlinear mapping from the original feature space to the embedding space whose dimension is $p=18$. The nonlinear mapping is used to predict the embedding of images which do not participate in the relative similarity comparisons. We use Regularized Least Square and Radial basis function kernel to obtain the nonlinear mapping.  The test set consists of $720$ images randomly chose from $18$ categories with $40$ images in each category.  We use ground truth category labels of training images to generate the triple-wise comparisons without any error. The ordinal constraints are generated like \cite{7410580}: if two images $i,\ j$ are from the same category and image $k$ is from the other categories, the similarity between $i$ and $j$ is larger than the similarity between $i$ and $k$, which is indicated by a triplet $(i,j,k)$. The total number of such triplets is $70,000$. Errors are then synthesized to simulate the human error in crowd-sourcing. We randomly sample $5\%$ and $10\%$ triplets to exchange the positions of $j$ and $k$ in each triplet $(i,j,k)$.
\\\\
\textbf{Evaluation Metrics.} To measure the effectiveness of various ordinal embedding methods for visual search, we consider three evaluation metrics, \textit{i.e.}, precision at top-K positions (Precision@K), recall at top-K positions (Recall@K), and Mean Average Precision (MAP). Given the mapping feature $\mathbf{X}=\{\mathbf{x}_1, \mathbf{x}_2,\dots,\mathbf{x}_{720}\}$ of test images and chosen an image $i$ belonging to class $c_i$ as a query, we sort the other images according to the distances between their embeddings and $\mathbf{x}_i$ in an ascending order as $\mathcal{R}_i$. True positives ($\text{TP}^K_i$) are images correctly labeled as positives, which involve the images belonging to $c_i$ and list within the top K in $\mathcal{R}_i$. False positives ($\text{FP}^K_i$) refer to negative examples incorrectly labeled as positives, which are the images belonging to $c_l(l\neq i)$ and list within the top K in $\mathcal{R}_i$. True negatives ($\text{TN}^K_i$) correspond to negatives correctly labeled as negatives, which refer to the images belonging to $c_l(l\neq i)$ and list after the top K in $\mathcal{R}_i$. Finally, false negatives ($\text{FN}^K_i$) refer to positive examples incorrectly labeled as negatives, which are relevant to the images belonging to class $c_i$ and list after the top K in $\mathcal{R}_i$. We are able to define Precision@K and Recall@K used in this paper as:
$
	\text{Precision@}K = \frac{1}{n}\sum_i^{n}p_i^K=\frac{1}{n}\sum_i^{n}\frac{\text{TP}^K_i}{\text{TP}^K_i+\text{FP}^K_i}
$
and
$
	\text{Recall@}K = \frac{1}{n}\sum_i^{n}r_i^K=\frac{1}{n}\sum_i^{n}\frac{\text{TP}^K_i}{\text{TP}^K_i+\text{FN}^K_i}.
$
Precision and recall are single-valued metrics based on the whole ranking list of images determined by the Euclidean distances among the embedding $\{\mathbf{x}_i\}_{i=1}^n$. It is desirable to also consider the order in which the images from the same category are embedded. By computing precision and recall at every position in the ranked sequence of the images for query $i$, one can plot a precision-recall curve, plotting precision $p_i(r)$ as a function of recall $r_i$. Average Precision(AP) computes the average value of $p_i(r)$ over the interval from $r_i=0$ to $r_i=1$: $\text{AP}_i = \int_{0}^1 p_i(r_i)dr_i$, which is the area under precision-recall curve. This integral can be replaced with a finite sum over every position $q$ in the ranked sequence of the embedding: $\text{AP}_i = \sum_{q=1}^{40} p_i(q)\cdot\triangle r_i(q)$, where $\triangle r_i(q)$ is the change in recall from items $q-1$ to $q$. The MAP used in this paper is defined as $\text{MAP} = \frac{1}{n}\sum_{i=1}^{n}\text{AP}_i$.
\\\\
\textbf{Results.} The experiment results are shown in Table \ref{tabl:3} and Figure \ref{fig:sun}.
As shown in Table \ref{tabl:3} and Figure \ref{fig:sun} with $K$ varying from $40$ to $100$, we observe that non-convex SVRG-SBB$_\epsilon$ consistently achieves the superior Precision@K, Recall@K and MAP results comparing against the other methods with the same gradient calculation. The results of GNMDS illustrate that SVRG-SBB$_\epsilon$ is more suitable for non-convex objective functions than the other methods. Therefore, SVRG-SBB$_\epsilon$ has a very promising potential in practice, because it generates appropriate step sizes automatically while running the algorithm and the result is robust. Moreover, under our setting and with small noise, all the ordinal embedding methods achieve the reasonable results for image retrieval. It illustrates that high-quality relative similarity comparisons can be used for learning meaningful representation of massive data, thereby making it easier to extract useful information in other applications.

\section{Conclusions}

In this paper, we propose a stochastic non-convex framework for the ordinal embedding problem. We propose a novel stochastic gradient descent algorithm called SVRG-SBB for solving this non-convex framework. The proposed SVRG-SBB is a variant of SVRG method incorporating with the so-called stabilized BB (SBB) step size, a new, stable and adaptive step size introduced in this paper. The main idea of the SBB step size is adding another positive term in the absolute value of the denominator of the original BB step size such that SVRG-SBB can overcome the instability of the original BB step size when applied to such non-convex problem. A series of simulations and real-world data experiments are implemented to demonstrate the effectiveness of the proposed SVRG-SBB for the ordinal embedding problem. It is surprising that the proposed SVRG-SBB outperforms most of the state-of-the-art methods in the perspective of both generalization error and computational cost. We also establish the $O(1/T)$ convergence rate of SVRG-SBB in terms of the convergence to a stationary point. Such convergence rate is comparable to the existing best convergence results of SVRG in literature.


\section*{Acknowledgment}
The work of Ke Ma is supported by National Key Research and Development Plan (No.2016YFB0800403), National Natural Science Foundation of China (No.U1605252 and 61733007). The work of Jinshan Zeng is supported in part by the National Natural Science Foundation of China (No.61603162), and the Doctoral start-up foundation of Jiangxi Normal University. The work of Qianqian Xu was supported in part by National Natural Science Foundation of China (No. 61672514), and CCF-Tencent Open Research Fund. Yuan Yao's work is supported in part by Hong Kong Research Grant Council (HKRGC) grant 16303817, National Basic Research Program of China (No. 2015CB85600, 2012CB825501), National Natural Science Foundation of China (No. 61370004, 11421110001), as well as grants from Tencent AI Lab, Si Family Foundation, Baidu BDI, and Microsoft Research-Asia. 

\bibliographystyle{aaai}
\bibliography{aaai18}

%
\twocolumn[\section*{Supplementary Materials for \\"Stochastic Non-convex Ordinal Embedding with Stabilized Barzilai-Borwein Step Size"}]

\section{Proof of Lemma \ref{lemma:1}}
	\label{proof:lemma1}
	\begin{proof}
		Let
		\begin{equation}
		\mathbf{X}_p =
		\begin{pmatrix}
		\mathbf{X}_1 \\
		\mathbf{X}_2
		\end{pmatrix} =
		\begin{pmatrix}
		\mathbf{x}_i \\
		\mathbf{x}_j \\
		\mathbf{x}_l \\
		\mathbf{x}_k
		\end{pmatrix}
		\end{equation}
		where $\mathbf{X}_1=[\mathbf{x}_i^T,\mathbf{x}_j^T]^T$, $\mathbf{X}_2=[\mathbf{x}_l^T,\mathbf{x}_k^T]^T$ and
		\begin{equation}
		\mathbf{M} =
		\left(\begin{array}{cc}
		\mathbf{M}_1 & \\
		& \mathbf{M}_2 \\
		\end{array}\right) =
		\left(\begin{array}{rrrr}
		\mathbf{I}& -\mathbf{I}&  &   \\
		-\mathbf{I}& \mathbf{I}&  &   \\
		&  & -\mathbf{I}& \mathbf{I} \\
		&  &  \mathbf{I}& -\mathbf{I}
		\end{array}\right)
		\end{equation}
		where $\mathbf{I}_{d\times d}$ is the identity matrix.
		
		The first-order gradient of CKL is
		\begin{equation}
		\begin{aligned}
		& \nabla f^{\text{ckl}}_p(\mathbf{X}) &=&\
		\frac{2\mathbf{MX}_p}{d^2_{ij}(\mathbf{X})+d^2_{lk}(\mathbf{X})+2\delta}\\
		& &-& \frac{2\left(\begin{array}{cc}\ \mathbf{O} & \\ & \mathbf{M}_2\mathbf{X}_2 \\ \end{array}\right)}{d^2_{lk}(\mathbf{X})+\delta}
		\end{aligned}
		\end{equation}
		and the second-order Hessian matrix of CKL is
		\begin{equation}
		\begin{aligned}
		& \nabla^2 f^{\text{ckl}}_p(\mathbf{X}) &=&\ \ \frac{2\mathbf{M}}{d^2_{ij}(\mathbf{X})+d^2_{lk}(\mathbf{X})+2\delta}\\
		& &-&\ \frac{4\mathbf{MX}_p\mathbf{X}_p^T\mathbf{M}}{[d^2_{ij}(\mathbf{X})+d^2_{lk}(\mathbf{X})+2\delta]^2} \\
		& &-&\ \frac{2\left(\begin{array}{cc}\ \mathbf{O} & \\ & \mathbf{M}_2 \\ \end{array}\right)}{d^2_{lk}(\mathbf{X})+\delta}\\
		& &+&\ \frac{4\left(\begin{array}{cc}\ \mathbf{O} & \\ & \mathbf{M}_2\mathbf{X}_2\mathbf{X}^T_2\mathbf{M}_2 \\ \end{array}\right)}{[d^2_{lk}(\mathbf{X})+\delta]^2}.
		\end{aligned}
		\end{equation}
		As $\mathbf{X}\in\mathbb{R}^{n\times d}$ is bounded, $\forall p=(i,j,l,k)$, $d^2_{ij}(\mathbf{X})$ and $d^2_{lk}(\mathbf{X})$ is bounded. So any element of $\nabla^2 f^{\text{ckl}}_p(\mathbf{X})$ is bounded , and $\nabla^2 f^{\text{ckl}}_p(\mathbf{X})$ has bounded eigenvalues. By the definition of Lipschitz continuity, we have the loss function of CKL (\ref{eq:scale-invariant}) has Lipschitz continuous gradient with bounded $\mathbf{X}$.
		
		\begin{figure*}[thb!]
			\begin{equation}
			\label{eq:gradient:student}
			\begin{aligned}
			& \nabla f^{\text{tste}}_p(\mathbf{X})=\frac{\alpha+1}{s_{ij}}\left(\begin{array}{cc}\mathbf{M}_1\mathbf{X}_1 & \\ & \mathbf{O} \\ \end{array}\right)-\frac{\alpha^{-\alpha}(\alpha+1)}{s_{ij}^{-\frac{\alpha+1}{2}}+s_{lk}^{-\frac{\alpha+1}{2}}}\left(\begin{array}{cc}s_{ij}^{-\frac{\alpha+3}{2}}\mathbf{M}_1\mathbf{X}_1 & \\ & s_{lk}^{-\frac{\alpha+3}{2}}\mathbf{M}_2\mathbf{X}_2 \\ \end{array}\right)
			\end{aligned}
			\end{equation}
		\end{figure*}
		
		\begin{figure*}[thb!]
			\begin{equation}
			\label{eq:hessian:student}
			\begin{aligned}
			& \nabla^2 f^{\text{tste}}_p(\mathbf{X}) &=&\ \ \ \frac{\alpha+1}{s_{ij}^2}\left(\begin{array}{cc}s_{ij}\mathbf{M}_1-2\mathbf{M}_1\mathbf{X}_1\mathbf{X}^T_1\mathbf{M}_1 & \\ & \mathbf{O} \\ \end{array}\right)-\frac{\alpha^{-\alpha}(\alpha+1)}{s_{ij}^{-\frac{\alpha+1}{2}}+s_{lk}^{-\frac{\alpha+1}{2}}}\left(\begin{array}{cc}s_{ij}^{-\frac{\alpha+3}{2}}\mathbf{M}_1 & \\ & s_{lk}^{-\frac{\alpha+3}{2}}\mathbf{M}_2\end{array}\right)\\
			& &+&\ \ \ \frac{\alpha^{-\alpha}(\alpha+1)^2}{\left(s_{ij}^{-\frac{\alpha+1}{2}}+s_{lk}^{-\frac{\alpha+1}{2}}\right)^2}\left(\begin{array}{cc}s_{ij}^{-(\alpha+3)}\mathbf{M}_1\mathbf{X}_1\mathbf{X}^T_1\mathbf{M}_1 & \\ & s_{lk}^{-(\alpha+3)}\mathbf{M}_2\mathbf{X}_2\mathbf{X}^T_2\mathbf{M}_2\end{array}\right)\\
			& &+&\ \ \ \frac{\alpha^{-\alpha}(\alpha+1)(\alpha+3)}{s_{ij}^{-\frac{\alpha+1}{2}}+s_{lk}^{-\frac{\alpha+1}{2}}}\left(\begin{array}{cc}s_{ij}^{-\frac{\alpha+5}{2}}\mathbf{M}_1\mathbf{X}_1\mathbf{X}_1^T\mathbf{M}_1 & \\ & s_{lk}^{-\frac{\alpha+5}{2}}\mathbf{M}_2\mathbf{X}_2\mathbf{X}_2^T\mathbf{M}_2\end{array}\right)
			\end{aligned}
			\end{equation}
		\end{figure*}
		
		The first-order gradient of STE (\ref{eq:logistic}) is
		\begin{equation}
		\nabla f^{\text{ste}}_p(\mathbf{X})=
		2\frac{\exp(d^2_{ij}(\mathbf{X})-d^2_{lk}(\mathbf{X}))}{1+\exp(d^2_{ij}(\mathbf{X})-d^2_{lk}(\mathbf{X}))}\mathbf{MX}_p
		\end{equation}
		and the second-order Hessian matrix of STE (\ref{eq:logistic}) is
		\begin{equation}
		\begin{aligned}
		& & & \nabla^2 f^{\text{ste}}_p(\mathbf{X})\\
		& &=&\ 4\frac{\exp(d^2_{ij}(\mathbf{X})-d^2_{lk}(\mathbf{X}))}{[1+\exp(d^2_{ij}(\mathbf{X})-d^2_{lk}(\mathbf{X}))]^2}\mathbf{MX}_p\mathbf{X}^T_p\mathbf{M}\\
		& &+&\ 2\frac{\exp(d^2_{ij}(\mathbf{X})-d^2_{lk}(\mathbf{X}))}{1+\exp(d^2_{ij}(\mathbf{X})-d^2_{lk}(\mathbf{X}))}\mathbf{M}.
		\end{aligned}
		\end{equation}
		All elements of $\nabla^2 f^{\text{ckl}}_p(\mathbf{X})$ are bounded. So the eigenvalues of $\nabla^2 f^{\text{ckl}}_p(\mathbf{X})$ are bounded. By the definition of Lipschitz continuity, (\ref{eq:logistic}) has Lipschitz continuous gradient with bounded $\mathbf{X}$.
		
		The first-order gradient of GNMDS (\ref{eq:hinge}) is
		\begin{equation}
		\nabla f^{\text{gnmds}}_p(\mathbf{X})=
		\left\{\begin{array}{cl}
		\mathbf{0}, &\ \text{if}\ d^2_{ij}(\mathbf{X})+1-d^2_{lk}(\mathbf{X})<0,\\
		2\mathbf{MX}_p, &\ \text{otherwise},
		\end{array}\right.
		\end{equation}
		and the second-order Hessian matrix of GNMDS (\ref{eq:hinge}) is
		\begin{equation}
		\nabla^2 f^{\text{gnmds}}_p(\mathbf{X})=
		\left\{\begin{array}{cl}
		\mathbf{0}, & \text{if}\ d^2_{ij}(\mathbf{X})+1-d^2_{lk}(\mathbf{X})<0,\\
		2\mathbf{M}, &\ \text{otherwise}.
		\end{array}\right.
		\end{equation}
		If $d^2_{ij}(\mathbf{X})+1-d^2_{lk}(\mathbf{X})\neq 0$ for all $p\in\mathcal{P}$, $\nabla f^{\text{gnmds}}_p(\mathbf{X})$ is continuous on $\{\mathbf{x}_i, \mathbf{x}_j, \mathbf{x}_l, \mathbf{x}_k\}$ and the Hessian matrix $\nabla^2 f^{\text{gnmds}}_p(\mathbf{X})$ has bounded eigenvalues. So GNMDS has Lipschitz continuous gradient in some quadruple set as $\{\mathbf{x}_i, \mathbf{x}_j, \mathbf{x}_l, \mathbf{x}_k\}\subset\mathbb{R}^{p \times 4}$. As the special case of $p=(i,j,l,k)$ which satisfied $d^2_{ij}(\mathbf{X})+1-d^2_{lk}(\mathbf{X})=0$ is exceedingly rare, we split the embedding $\mathbf{X}$ into pieces $\{\mathbf{x}_i, \mathbf{x}_j, \mathbf{x}_l, \mathbf{x}_k\}$ and employ SGD and SVRG to optimize the objective function of GNMDS (\ref{eq:hinge}). The empirical results are showed in the Experiment section.
		
		Note $s_{ij} = \alpha+d^2_{ij}(\mathbf{X})$, and the first-order gradient of TSTE is (\ref{eq:gradient:student}).
		The second-order Hessian matrix of TSTE is (\ref{eq:hessian:student}). The boundedness of eigenvalues of The loss function of $\nabla^2 f^{\text{tste}}_p(\mathbf{X})$ can infer that the TSTE loss function (\ref{eq:student}) has Lipschitz continuous gradient with bounded $\mathbf{X}$.
		
		We focus on  a special case of quadruple comparisons as $i=l$ and $\{i,j,i,k\}\subset[n]^3$ in the Experiment section. To verify the Lipschitz continuous gradient of ordinal embedding objective functions with $c=(i,j,i,k)$ as $i=l$, we introduction the matrix $\mathbf{A}$ as
		\begin{equation}
		\mathbf{A} =
		\left(\begin{array}{rrrr}
		\mathbf{I} & \mathbf{0} & \mathbf{0} & \mathbf{0}\\
		\mathbf{0} & \mathbf{I} & \mathbf{0} & \mathbf{I}\\
		\mathbf{0} & \mathbf{0} & \mathbf{0} & \mathbf{I}
		\end{array}\right).
		\end{equation}
		By chain rule for computing the derivative, we have
		\begin{equation}
		\begin{aligned}
		& \nabla f_{ijk}(\mathbf{X}) &=&\ \ \ \mathbf{A}\nabla f_{ijlk}(\mathbf{X}),\\
		& \nabla^2 f_{ijk}(\mathbf{X}) &=&\ \ \mathbf{A}\nabla^2 f_{ijlk}(\mathbf{X})\mathbf{A}^T.
		\end{aligned}
		\end{equation}
		where $l = i$. As $\mathbf{A}$ is a constant matrix and $\nabla^2 f_{ijlk}(\mathbf{X})$ is bounded, all elements of the Hessian matrix $\nabla^2 f_{ijk}(\mathbf{X})$ are bounded. So the eigenvalues of $\nabla^2 f_{ijk}(\mathbf{X})$ is also bounded. The ordianl embedding functions of CKL, STE and TSTE with triplewise compsrisons have Lipschitz continuous gradient with bounded $\mathbf{X}$.
	\end{proof}

\section{Proof of Theorem \ref{svrg_bb_nonconvex}}

In this section, we prove our main theorem, i.e., Theorem \ref{svrg_bb_nonconvex}. Before doing this, we need some lemmas.

\begin{lemma}
\label{lemma:analytic}
Given some positive integer $l\geq 2$, then for any $0<x<\frac{1}{l}$, the following holds
\[
(1+x)^l \leq e^{l x} \leq 1 + 2 l x.
\]
\end{lemma}
\begin{proof}
Note that
\[(1+x)^l = e^{l\cdot \ln(1+x)} \leq e^{lx},\]
where the last inequality holds for $\ln(1+x) \leq x$ for any $0<x<1$. Thus, we get the first inequality. Let $h(z) = 1+2z-e^z$ for any $z\in (0,1)$. It is easy to check that $h(z) \geq 0$ for any $z \in (0,1)$. Thus we get the second inequality.
\end{proof}

In the following, we provide a key lemma that shows the convergence behavior of the inner loop of the proposed algorithm.
Before presenting it, we define several constants and sequences. For any $0\leq s \leq S-1$, let
\begin{align}
&\beta_s : = 4(m-1)L^3 \eta_s^2, \label{eq:batas}\\
&\rho_s : = 1+2\eta_s^2L^2[2(m-1)\eta_sL+1]\label{eq:rhos}.
\end{align}
Let $\{c_t^{s+1}\}_{t=1}^m$ be a nonnegative sequence satisfying $c_m^{s+1}=0$ and for $ t = m-1,\ldots,1$,
\[
c_t^{s+1} = c_{t+1}^{s+1} \left( 1+\eta_s \beta_s + 2\eta_s^2 L^2\right) + \eta_s^2 L^3.
\]
It is obvious that $\{c_t^{s+1}\}_{t=1}^m$ is monotonically decreasing as $t$ increasing, and
\[
c_1^{s+1} = \frac{\eta_s L^3 \cdot (\rho_s^{m-1}-1)}{\beta_s + 2\eta_s L^2}.
\]
Then we define
\begin{align}
\label{eq:Gammas}
\Gamma_t^{s+1} : = \eta_s \left[ 1-\frac{c_{t+1}^{s+1}}{\beta_s} - \eta_s (1+2c_{t+1}^{s+1})\right],
\end{align}
for any $0\leq t \leq m-1$ and $0 \leq  s \leq S-1$.

With the help of the sequences defined above, we present a key lemma as follows.
\begin{lemma}[Convergence behavior of inner loop]
\label{keylemma}
Let $\{{\bf X}_t^{s+1}\}_{t=1}^m$ be a sequence generated by Algorithm \ref{alg:svrg-bb} at the $s$-th inner loop, $s=0,\ldots, S-1$. If the following conditions hold
\begin{align}
& \eta_s + 4(m-1)\eta_s^3 L^3 < 1/2, \label{cond1}\\
& (m-1)\eta_s^2 L^2\left[ 2(m-1)\eta_s L +1\right] <1/2 \label{cond2},
\end{align}
then
\[
\mathbb{E}[\|\nabla F({\bf X}_t^{s+1})\|^2] \leq \frac{R_t^{s+1} - R_{t+1}^{s+1}}{\Gamma_t^{s+1}},
\]
where $R_t^{s+1}:= \mathbb{E}[F({\bf X}_t^{s+1}) + c_t^{s+1} \|{\bf X}_t^{s+1} - \tilde{\bf X}^s\|^2]$, $0\leq t \leq m-1.$
\end{lemma}
\begin{proof}
This lemma is a special case of \cite[Lemma 1]{pmlr-v48-reddi16} with some specific choices of $\beta_s$. Thus, we only need to check the defined $\Gamma_t^{s+1}$ is positive for any $0\leq t\leq m-1$. In order to do this, we firs check that
\begin{align}
\label{c-beta}
c_t^{s+1} < \frac{\beta_s}{2}, \quad t=1,\ldots,m
\end{align}
under the condition \eqref{cond2} of this lemma.

Since $c_t^{s+1}$ is monotonically decreasing, thus \eqref{c-beta} is equivalent to showing $c_1^{s+1} < \frac{\beta_s}{2}$, which implies
\begin{align}
\label{c-beta1}
\eta_sL^3(\rho_s^{m-1}-1) < \frac{1}{2}\beta_s(\beta_s + 2\eta_s L^2).
\end{align}
By Lemma \ref{lemma:analytic}, if \eqref{cond2} holds, then
\[
\rho_s^{m-1} < 1+4(m-1)\eta_s^2 L^2 \left[ 2(m-1)\eta_s L+1\right].
\]
It implies
\begin{align*}
\eta_s L^3 (\rho_s^{m-1}-1)
& < 4(m-1)\eta_s^3 L^5 \left[ 2(m-1)\eta_s L +1 \right]\\
& = 2(m-1)\eta_s^2 L^3 \left[ 4(m-1)\eta_s^2 L^3 + 2\eta_s L^2 \right]\\
& = \frac{1}{2}\beta_s (\beta_s + 2\eta_s L^2).
\end{align*}
Thus, \eqref{c-beta} holds.

In the next, we prove $\Gamma_t^{s+1}>0$, $t=0,\ldots, m-1$. By the definition of $\Gamma_t^{s+1}$ \eqref{eq:Gammas}, it holds
\begin{align*}
\Gamma_t^{s+1}
& \geq \eta_s \left[ 1 - \frac{c_1^{s+1}}{\beta_s} - \eta_s (1+2c_1^{s+1})\right]\\
&> \eta_s \left[ \frac{1}{2} - \eta_s (1+\beta_s)\right],
\end{align*}
where the last inequality holds for \eqref{c-beta}.
Thus, if \eqref{cond1} holds, then obviously, $\Gamma_t^{s+1}>0$, $t=0,\ldots,m-1.$
Therefore, we end the proof of this lemma.
\end{proof}

Based on the above lemma and the existing \cite[Theorem 2]{pmlr-v48-reddi16}, we directly have the following proposition on the convergence of SVRG with a generic step size $\{\eta_s\}_{s=0}^{S-1}$.

\begin{proposition}
\label{svrg-generic}
Let $\{\{{\bf X}_t^s\}_{t=1}^m\}_{s=1}^S$ be a sequence generated by SVRG with a sequence of generic step sizes $\{\eta_s\}_{s=0}^S$. If the following conditions hold:
\begin{align}
&\eta_s + 4(m-1)\eta_s^3 L^3 < 1/2,\tag{\ref{cond1}}\\
&(m-1)\eta_s^2L^2\left[ 2(m-1)\eta_s L +1 \right]<1/2,\tag{\ref{cond2}}
\end{align}
then we have
\[
\mathbb{E}[\|F({\bf X}_{\mathrm{out}})\|^2] \leq \frac{F(\tilde{\bf X}^0) - F({\bf X}^*)}{m \cdot S \cdot \gamma_S},
\]
where ${\bf X}^*$ is an optimal solution to \eqref{eq:2}, and
\begin{align*}
\gamma_S
&: = \min_{0\leq s \leq S-1,\ 0 \leq t \leq m-1} \Gamma_t^{s+1}\\
&\geq \min_{0\leq s \leq S-1} \left\{\eta_s\left[ \frac{1}{2} - \eta_s(1+4(m-1)L^3\eta_s^2)\right]\right\}.
\end{align*}
\end{proposition}

Based on Proposition \ref{svrg-generic}, we show the proof of Theorem \ref{svrg_bb_nonconvex}.

\begin{proof}[Proof of Theorem \ref{svrg_bb_nonconvex}]
By \eqref{eq:bound-rbb}, we have
\[
\eta_s \leq \frac{1}{m\epsilon}, \quad s = 0,\ldots, m-1.
\]
In order to show the conditions \eqref{cond1}-\eqref{cond2} in Proposition \ref{svrg-generic}, it suffices to show
\begin{align*}
&\frac{1}{m\epsilon} + 4m\cdot L^2 \cdot \frac{1}{m^3\epsilon^3} < 1/2,\\
& m \cdot \frac{1}{m^2\epsilon^2} \cdot L^2 \left[2m \cdot \frac{1}{m \epsilon} +1 \right]<1/2.
\end{align*}
Thus, the convergence condition on $m$ in Theorem \ref{svrg_bb_nonconvex} can be easily derived by solving the above two inequalities.

The second part of Theorem \ref{svrg_bb_nonconvex} for the SBB$_0$ step size holds by noting that in this case, the upper bound of $\eta_s$
\[
\eta_s \leq \frac{1}{m\mu}, \quad s = 0,\ldots, m-1.
\]
The convergence rate can be derived by the similar proof of the first part.
Therefore, we end the proof of this theorem.
\end{proof}

\end{document}